\pdfoutput=1
\documentclass[letterpaper, 10 pt, conference]{ieeeconf}  

\usepackage{blindtext, graphicx}
\usepackage{gensymb}
\usepackage{xcolor}

\usepackage{tikz,lipsum}
\usepackage[labelformat=simple]{subcaption}

\DeclareCaptionLabelSeparator{periodspace}{.\quad}
\usepackage{listings}

\usepackage{algorithm}
\usepackage{algorithmic}

\usepackage{ amssymb }
\usepackage{amsmath}
\usepackage{commath}

\usepackage{framed} 

\usepackage{float}

\newcommand*{\prob}{\mathsf{P}}

\IEEEoverridecommandlockouts                              

\title{\LARGE \bf
Robot Localisation and 3D Position Estimation Using a Free-Moving Camera and Cascaded Convolutional Neural Networks
}

\author{Justinas Mi\v{s}eikis$^{1}$, Patrick Kn\"obelreiter$^{2}$, Inka Brijacak$^{3}$, Saeed Yahyanejad$^{4}$, Kyrre Glette$^{5}$, Ole Jakob Elle$^{6}$, \\Jim Torresen$^{7}$
\thanks{$^{1}$ $^{5}$ $^{6}$ $^{7}$Justinas Mi\v{s}eikis, Kyrre Glette, Ole Jakob Elle and Jim Torresen are with the Department of Informatics, University of Oslo, Oslo, Norway}
\thanks{$^{2}$ Patrick Kn\"obelreiter is with the Institute of Computer Graphics and Vision, Graz University of Technology, Graz, Austria}
\thanks{$^{3}$ $^{4}$ Inka Brijacak and Saeed Yahyanejad are with the Joanneum Research - Robotics, Klagenfurt am W\"orthersee, Austria} 
\thanks{$^{6}$Ole Jakob Elle has his main affiliation with The Intervention Centre, Oslo University Hospital, Oslo, Norway {\tt\small oelle@ous-hf.no}}%
\thanks{$^{1}$ $^{5}$ $^{7}$ {\tt\small \{justinm,kyrrehg,jimtoer\}@ifi.uio.no}}%
\thanks{$^{2}$ {\tt\small knoebelreiter@icg.tugraz.at}}%
\thanks{$^{3}$ {\tt\small Inka.Brijacak@joanneum.at}}%
\thanks{$^{4}$ {\tt\small Saeed.Yahyanejad@joanneum.at}}%
}

\begin{document}

\maketitle
\thispagestyle{empty}
\pagestyle{empty}

\begin{abstract}

Many works in collaborative robotics and human-robot interaction focuses on identifying and predicting human behaviour while considering the information about the robot itself as given. This can be the case when sensors and the robot are calibrated in relation to each other and often the reconfiguration of the system is not possible, or extra manual work is required. We present a deep learning based approach to remove the constraint of having the need for the robot and the vision sensor to be fixed and calibrated in relation to each other. The system learns the visual cues of the robot body and is able to localise it, as well as estimate the position of robot joints in 3D space by just using a 2D color image. The method uses a cascaded convolutional neural network, and we present the structure of the network, describe our own collected dataset, explain the network training and achieved results. A fully trained system shows promising results in providing an accurate mask of where the robot is located and a good estimate of its joints positions in 3D. The accuracy is not good enough for visual servoing applications yet, however, it can be sufficient for general safety and some collaborative tasks not requiring very high precision. The main benefit of our method is the possibility of the vision sensor to move freely. This allows it to be mounted on moving objects, for example, a body of the person or a mobile robot working in the same environment as the robots are operating in.

\end{abstract}



\section{INTRODUCTION}

Robotic manipulators are becoming cheaper resulting in new application fields outside the traditional industrial environment. It is more common to see robots in hospitals, warehouses and households. These environments result in robots having to share the workspace with people and even perform collaborative tasks. The concept of a shared workspace has been an active research area for many years, which is still highly relevant today~\cite{roach1987coordinating}~\cite{leitner2012transferring}. Furthermore, with the appearance of Industry 4.0, the need toward the environment and safety-aware robots is growing~\cite{lee2015cyber}.


Collaborative robots, like Baxter and Sawyer, are advertised to be fully safe around people, however, it commonly means that they have sophisticated collision detection systems~\cite{fitzgerald2013developing}. Ideally, collisions should be avoided at all, especially in sensitive environments like hospitals. Collision avoidance can be achieved by adding vision sensors.

Vision sensors observe the environment and indicate the areas which are unobstructed and safe to operate in, and are used to plan the robot movements accordingly. However, there are issues with this approach. Sensors have to be fixed on the robot itself or fixed in relation to the robot body. A precise Hand-Eye calibration is then performed to allow the sensors and the robot to operate in the same coordinate system. However, then the setup takes up more space and any unexpected disturbances or repositioning of the sensor can mean that the calibration has to be repeated. Despite automated calibration procedures, the process can still be cumbersome and time consuming~\cite{miseikis2016automatic}. Another option would be to fix the vision sensor on the robot body itself, commonly on the end-effector of the robot. This can be an effective method for collision avoidance for the end-effector of the robot, however, the field-of-view is normally limited and a full robot body collision check is rarely possible~\cite{flandin2000eye}.


\begin{figure*}[ht]
\vspace{0.2cm}
\centering
    \includegraphics[width=0.99\linewidth]{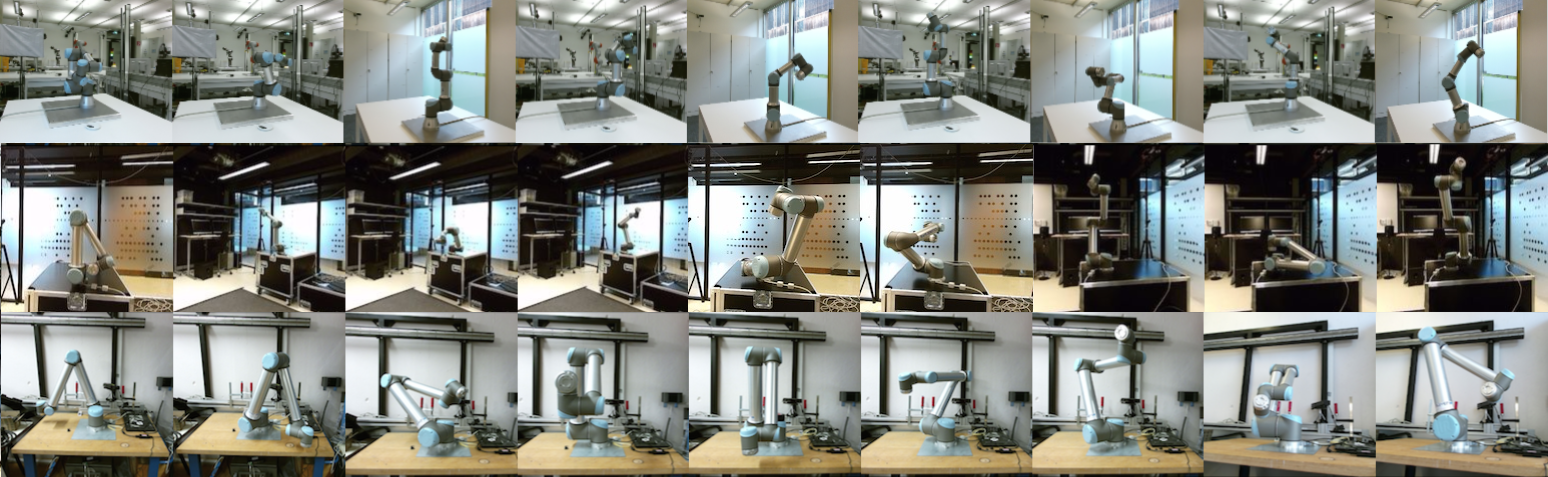}
    \caption{Samples from a collected robot dataset. Each row of images represents different robot type in the following order: UR3, UR5 and UR10. The dataset was created using a varying background to provide more robustness.}
    \label{fig:dataset_example_images}
\end{figure*}

There has been a significant amount of work towards robot autonomy and self-localisation. However, robot self-awareness is normally limited to navigation, especially for mobile robots, or self-collision avoidance for robot arms or humanoid robots, where the robot model is known~\cite{schneegans2007using}~\cite{gutmann1996amos}~\cite{stasse2008real}~\cite{de2007skeleton}.

Visual-based robot manipulator tracking has been extensively researched as well. End-effector being the main point of focus with the aim of conducting robot control based on visual servoing~\cite{wilson1996relative}~\cite{ruf1997visual}. Furthermore, it has proven to be an effective method for adaptive redundant robot control in Cartesian space~\cite{daachi2006neural}. Image-based tracking of 7-DoF robot arm showed promising results with dynamic parameter tuning as well~\cite{siradjuddin2014image}. Interesting work was presented, where authors are using particle swarm optimisation approach for fuzzy sliding mode control to track the end-effector of the robot manipulator~\cite{soltanpour2013particle}. Despite achieving good accuracy, most of these methods used prior knowledge or fixed setups for the particular use case. Changing the setup would result in the need to tune the algorithm for it to function accurately given the new conditions. Furthermore, commonly it was just the end-effector of the robot that was tracked instead of the whole robot body.

When looking at the field of human-robot interaction, a significant amount of work has been done on the design of the systems and workspaces allowing to monitor the human presence in close proximity to the robot and detect any irregularities~\cite{michalos2015design}~\cite{sheridan2016human}~\cite{brijacak2017}. Another work is focusing on the best approaches to safeguard the workspace of the robots~\cite{yamada1997human}.

When looking at the motion planning and behaviour prediction topics, most of the focus has been on modelling the human motions~\cite{mainprice2013human}. A heatmap of the workspace could be constructed to allow the robot to predict where dynamic obstacles are most likely to enter and have an additional reflexive behaviour override for unexpected cases~\cite{mivseikis2016multi}.

However, the majority of existing research has a robot model and control architectures well defined and fine-tuned. This means that the hardware setups are usually fixed and all the sensors have to be attached at the defined locations and calibrated specifically for the use case.

Our current research focuses on adding the flexibility on the robot identification side and allowing more unrestricted setups. For example, having a free-moving vision sensor as a part of the robotic system aimed at the robot safety or human-robot collaboration use case. We address this issue by removing the need for fixed setups. Instead of having a known transformation matrix between the coordinate frames of the sensor and the robot base, we teach the system to identify the robot body in a color image provided by the vision sensor. Our method uses convolutional neural networks (CNNs), which learn visual cues allowing it to understand the environment~\cite{simard2003best}. The system identifies the robot body in the color image of the vision sensor, and depth information normally provided by 3D cameras is not needed for the recognition task anymore. Furthermore, the system estimates the robot body configuration and 3D coordinates of each joint of the robot.

The vision sensor can be placed anywhere around the robot or even used as a wearable device by the robot operator. This approach can prove very useful in a cluttered environment where one or many robots are located, such as a factory floor or automated surgery room. Such environments can have a limited space for fixed camera setups or line-of-sight can be blocked by people or other machinery operating in close proximity. Having multiple free-moving cameras is one of the robust solutions ensuring the workspace is constantly observed. An operator or a visitor can have a wearable camera which observes the surroundings and indicates the positions of all the robots in the vicinity. A warning system or even an emergency stop option can be incorporated for the situations when the robot gets too close to the person to ensure a fully-safe operation.

Systems using our approach can also be useful in robot-robot interaction cases, where a mobile robot is operating in the same environment as robotic manipulators. It should avoid getting too close to other robots and avoid possible collisions. Even given a fully known environment, our system can prove useful if navigation or localisation algorithm fails to get an accurate position estimate, the vision sensor on the mobile robot can indicate positions where other robots are located. It can be useful for robot-to-robot interaction tasks. For example, if a mobile robot is bringing tools or objects that a robot arm needs to grasp, the mobile robot could localise itself in relation to the robot manipulator.

This paper is organized as follows. We present the system setup and dataset collection in Section~\ref{sec:system_setup}. Then,  we explain the proposed method and CNN architecture in Section~\ref{sec:method}. We provide experimental results in Section~\ref{sec:results}, followed by relevant conclusions and future work in Section~\ref{sec:conclusion}.

\begin{figure}[ht]
\centering
\begin{subfigure}[t]{0.23\textwidth}
    \includegraphics[width=\textwidth]{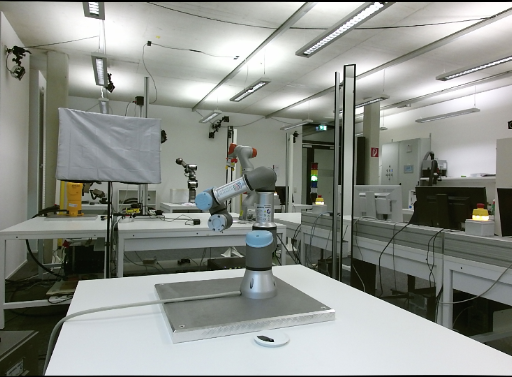}
    \caption{Color Image}
    \label{fig:input_color_image}
\end{subfigure}
~
\begin{subfigure}[t]{0.23\textwidth}
    \includegraphics[width=\textwidth]{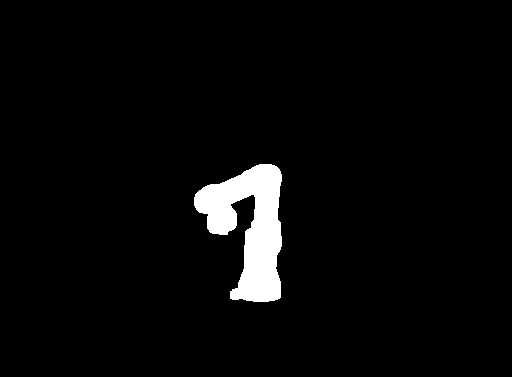}
    \caption{Ground truth model of the robot mask}
    \label{fig:input_gt_image}
\end{subfigure}
\caption{Example image of the dataset containing an UR3 robot.}
\label{fig:input_data}
\end{figure}


\section{SYSTEM SETUP AND DATASET COLLECTION}
\label{sec:system_setup}

In our experiments, we use three types of robot arms produced by Universal Robots: UR3, UR5 and UR10. All three robots have a similar appearance, but different size and payload capabilities. A Kinect V2 camera was used as a 3D vision sensor providing both color image and depth information, needed to create the dataset containing ground truth data~\cite{Fankhauser2015KinectV2ForMobileRobotNavigation}. The final, fully-trained system only needs the color image. The whole system was based on a combination of the Robot Operating System (ROS) and Theano framework~\cite{quigley2009ros}.

\begin{figure}[h]
    \includegraphics[width=0.48\textwidth]{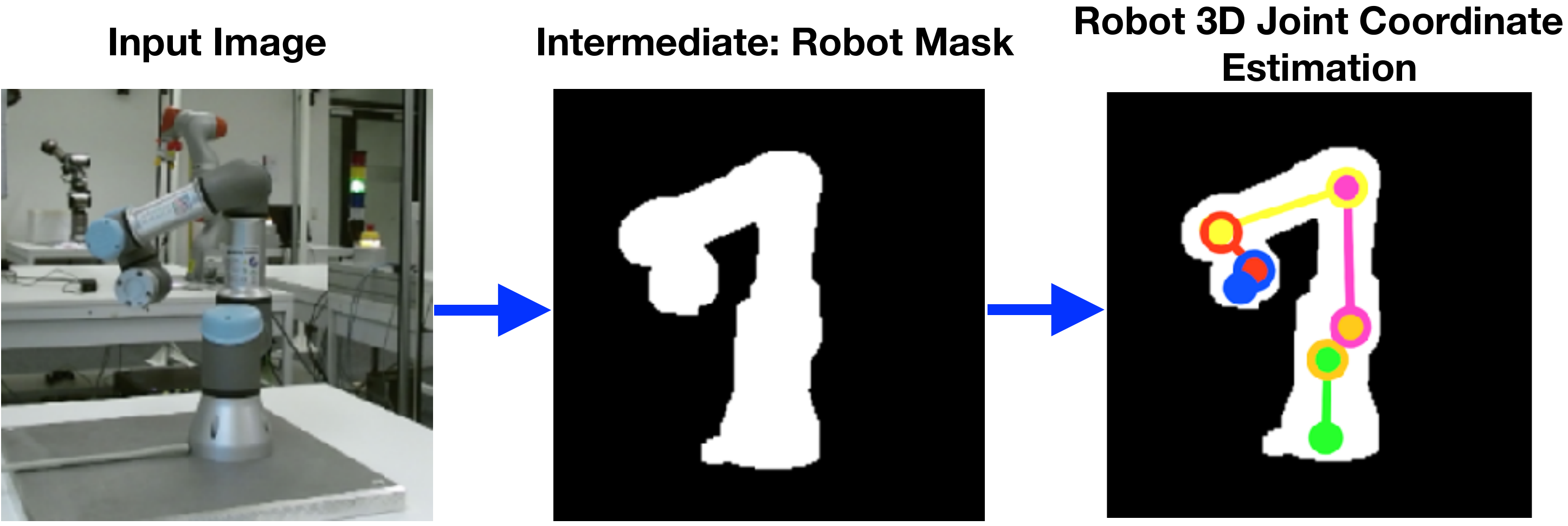}
    \caption{Process described in regards to inputs and outputs of the system. A simple color image of the robot body is provided as an input to the system. The first CNN estimates the mask containing the robot body and this result is overlayed with the color image and used as an input to the second CNN. The second CNN provides an estimate of the joint coordinates of the robot in 3D. Each robot joint is visualised with a circle of a different color.}
    \label{fig:cnn_process_images}
\end{figure}

Deep learning requires a large amount of diverse training data to ensure efficient and robust learning. Given a limited availability of datasets for such applications, it was decided to create a dataset for this purpose. Access to the robots was obtained in three institutions: TU Graz, Joanneum Research and the University of Oslo.

In order to obtain a precise ground truth data, a Kinect V2 camera was placed at a number of positions overlooking the robot. At each position, a precise Hand-Eye calibration was performed by placing a marker on the end-effector of the robot and using both color and depth image for the calibration process~\cite{heikkila2000flexible}. Having a precise coordinate system transformation from the camera to the robot base allows us to know precisely where the robot is located in the camera image.

\begin{table}[h]
\caption{Dataset summary describing the number of samples collected for each type of the robot. In total 9 recordings were made, 3 for each type of the robot.}
\label{table:dataset_summary}
\centering
\begin{tabular}{ |p{2.0cm}||p{1.5cm}|p{3cm}|}
 \hline
 Recording & Robot Type & Number of Samples \\
 \hline
 Rec 1 & UR3 & 211\\
 Rec 2 & UR3 & 252\\
 Rec 3 & UR3 & 463\\
 \hline
 Rec 4 & UR5 & 252\\
 Rec 5 & UR5 & 756\\
 Rec 6 & UR5 & 1512\\
 \hline
 Rec 7 & UR10 & 112\\
 Rec 8 & UR10 & 278\\
 Rec 9 & UR10 & 514\\
 \hline
\end{tabular}
\end{table}


We used the MoveIt! package~\cite{sucan2013moveit} to obtain ground truth data by using a robot self-filtering algorithm. At each time instance, the robot joints encoder information is combined with a simplified robot model, which is taken from the Unified Robot Description Format (URDF) file~\cite{meeussen2012urdf}, to generate a precise estimation of the current robot pose in 3D space and a robot body mask as a 2D image. It can be used to find the robot in both, color and depth image.

Robot movements were pre-programmed in joints coordinate system to move in as many different configurations as possible without hitting an obstacle or self-collision occurring. Each of the robot joints is moved through the full range of motion in combination with other joints as well. The step size of the joint movements is varied between the datasets resulting in a different number of samples in each. This method ensured that the robot body will be observed from many angles by the vision sensor. After each movement was performed, a trigger signal was sent to record camera images, joint coordinates, Cartesian coordinates of each joint and ground-truth model of the robot position. The number of samples per dataset varied from $112$ to $1512$. In total nine datasets were collected, three for each type of the robot, summarized in Table~\ref{table:dataset_summary}. Example images from the collected dataset can be seen in Figure~\ref{fig:dataset_example_images}. Datasets with the UR5 robot were the most extensive given the access to the robot at the lab of the main author. An example of color and ground truth images can be seen in Figure~\ref{fig:input_data}.

All the collected datasets were used for the training process, resulting in $926$ samples for UR3, $2520$ samples for UR5 and $904$ samples for UR10. The datasets were split into training and test set by randomly assigning 80$\%$ and 20$\%$ of the images accordingly. All of the images have the resolution of $512\times424$ pixels and are rectified using an internal camera calibration to remove the lens distortion. Higher resolution, $960\times540$ pixels color images were recorded as well, however, in practice, we scaled and cropped images to have the same resolution for all the types: color, depth and ground truth mask to avoid any scaling issues.

\section{METHOD}
\label{sec:method}

Our method is based on a two-level cascaded CNN, where one CNN is used for the classification task in foreground/background detection of the robot body in the image, and the second CNN is used for landmark detection of the robot joint positions in 3D coordinates. The process in regards to the input and output images is shown in Figure~\ref{fig:cnn_process_images}.

The principle of CNN is to have an image as an input, which is passed to the network. Normally, CNN contains a number of hidden layers, which lead to the output, which is also given during the training process, defined as ground truth. In the hidden layers, the network is capable of learning a number of filters, which help to achieve the desired result, thus minimising the error between the output of the network and provided ground truth result. The learning process is done by initialising random weights, getting the output, comparing it to the desired result and then adjusting the weights in the hidden layers during the back-propagation process in order to achieve better accuracy.


\begin{figure*}[ht]
\vspace{0.2cm}
    \hfill
    \centering
    \begin{subfigure}[t]{\textwidth}
        \centering
        \includegraphics[width=0.99\linewidth]{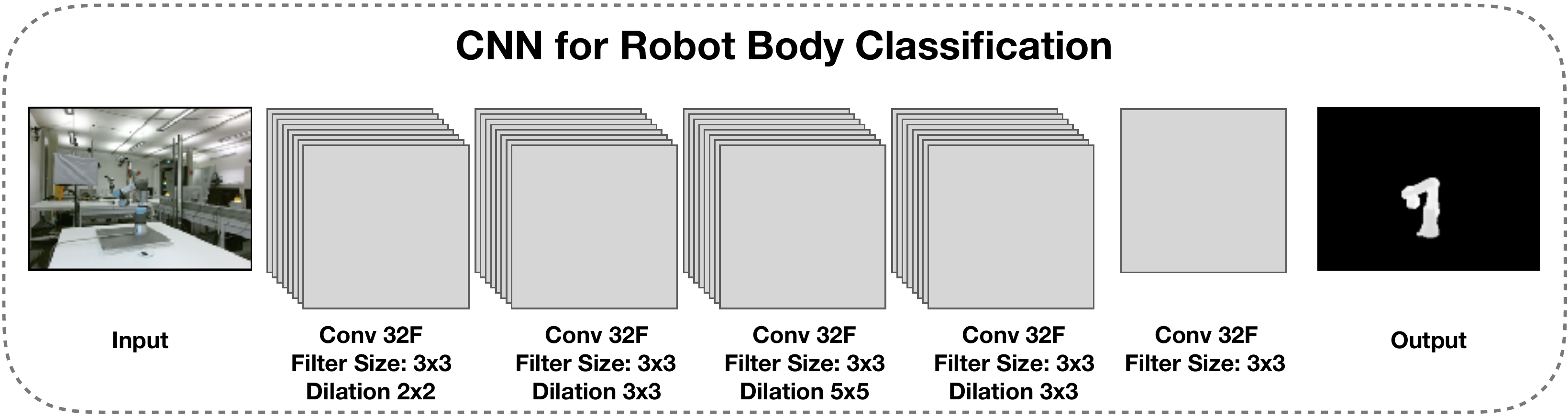}
        \caption{CNN architecture for the robot mask classification. The network consists of 5 convolutional layers with varying dilation. Input is a color image and output is a mask image defining the body of the robot.}
        \label{fig:cnn_classification} 
    \end{subfigure}
    \hfill
    \begin{subfigure}[t]{\textwidth}
        \centering
        \includegraphics[width=0.99\linewidth]{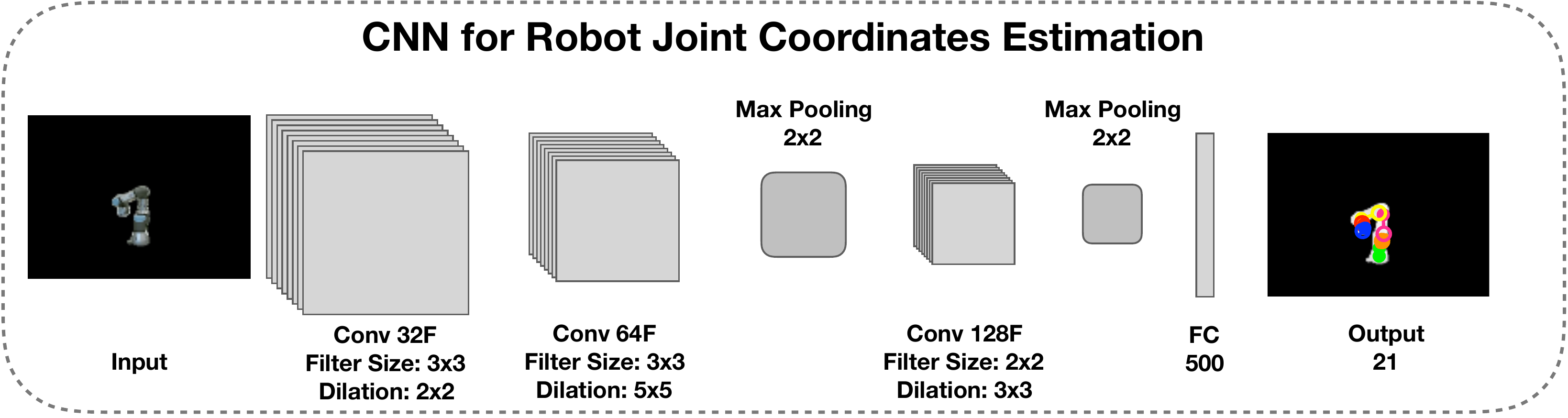}
        \caption{CNN architecture for the robot joint coordinate estimation. The network consists of 3 convolutional layers, 2 pooling layers and a fully connected layer in the end. Input is an overlayed color image with a robot foreground mask and output is 3D coordinates of the robot joints in the coordinate system of the vision sensor.}
        \label{fig:cnn_coords}
    \end{subfigure}
    
    \caption{Cascaded CNN architecture for the robot position estimation using non-fixed camera.}
    
\end{figure*}

For the robot body classification, our CNN consists of four convolutional layers with 32 filters each and varying dilation was used as well as the last convolutional layer containing just one filter. The details about the architecture are illustrated in Figure~\ref{fig:cnn_classification}. The loss function was specifically designed to take into consideration the rather small area of the foreground object in the input image. In most of the cases, the area of the robot body in the input image was $6-17\%$ of the whole image. Without this adjustment, in some cases, the network classifying all the pixels as background could still achieve the accuracy of over $90\%$, which is conceptually wrong. Calculation of the foreground weight $w_{fg}$ is shown in Equation~\ref{eq:fg_weight}. It is based on the inverse probability of the foreground and background classes, where $Y \in \{fg, bg\}$.

\begin{equation}
    w_{fg} = \frac{1}{\prob(Y=fg)}
\label{eq:fg_weight}
\end{equation}

Similarly, the background weight $w_{bg}$ is calculated using Equation~\ref{eq:bg_weight}.

\begin{equation}
    w_{bg} = \frac{1}{\prob(Y=bg)}
\label{eq:bg_weight}
\end{equation}


The loss function is calculated by first getting loss per pixel and then using it to calculate loss of the whole image. A normalisation factor $\mathcal{N}$, which is a number of pixels in the image, allows us to keep the learning rate fixed, independent of the image size. 


Loss function for one pixel $l^n$ is defined in Equation~\ref{eq:loss_classification_pixel}, where $i_{est}$ is $\prob(Y = fg)$, $(1-i_{est})$ is $\prob(Y = bg)$ and $i_{gt}$ is the ground truth value from the mask image.

\begin{equation} \label{eq:loss_classification_pixel}
    \begin{split}
        l^n (I_{est}^n, I_{gt}^n) = 
        & -w_{fg} i_{est} \log{(i_{gt})} \\
        & - w_{bg}(1-i_{est})\log{(1-i_{gt})}
    \end{split}
\end{equation}

The result is then used to calculate normalised loss for the whole image $\mathcal{L}_{mask}$ using Equation~\ref{eq:loss_classification_full}.
    
\begin{equation}  \label{eq:loss_classification_full}
    \mathcal{L}_{mask} (I_{est}, I_{gt}) = \frac{1}{\mathcal{N}} \sum\limits_{n} l^n (i_{est},  i_{gt})
\end{equation}

We formulate the joint coordinate estimation as a regression task using the second CNN. The network consists of three dilated convolutional layers with $32$, $64$ and $128$ filters respectively, two max-pooling layers in between and a fully connected layer in the end. Details of the network architecture can be seen in Figure~\ref{fig:cnn_coords}.

Loss function $\mathcal{L}_{coords}$ is based on Euclidean distance calculations between estimated and ground truth values as defined in Equation~\ref{eq:loss_coords}, where $N_j$ is the number of joints, $J_{i}$ defines ground truth position of each joint and $E_{i}$ is the estimated position of each joint by CNN.

\begin{equation} \label{eq:loss_coords}
    \mathcal{L}_{coords} = \frac{1}{N_j} \sum\limits_{i=1}^{N_j} \norm{J_{i}-E_{i}}_2
\end{equation}

In this work we decided not to use any prior robot model information to keep the system more adaptable to other robot models in the future, meaning that a raw CNN output is used to evaluate the accuracy of the results without any additional post-processing.

\begin{figure*}[ht]
\vspace{0.2cm}
    \hfill
    \centering
    \begin{subfigure}[t]{0.325\textwidth}
        \centering
        \includegraphics[width=\linewidth]{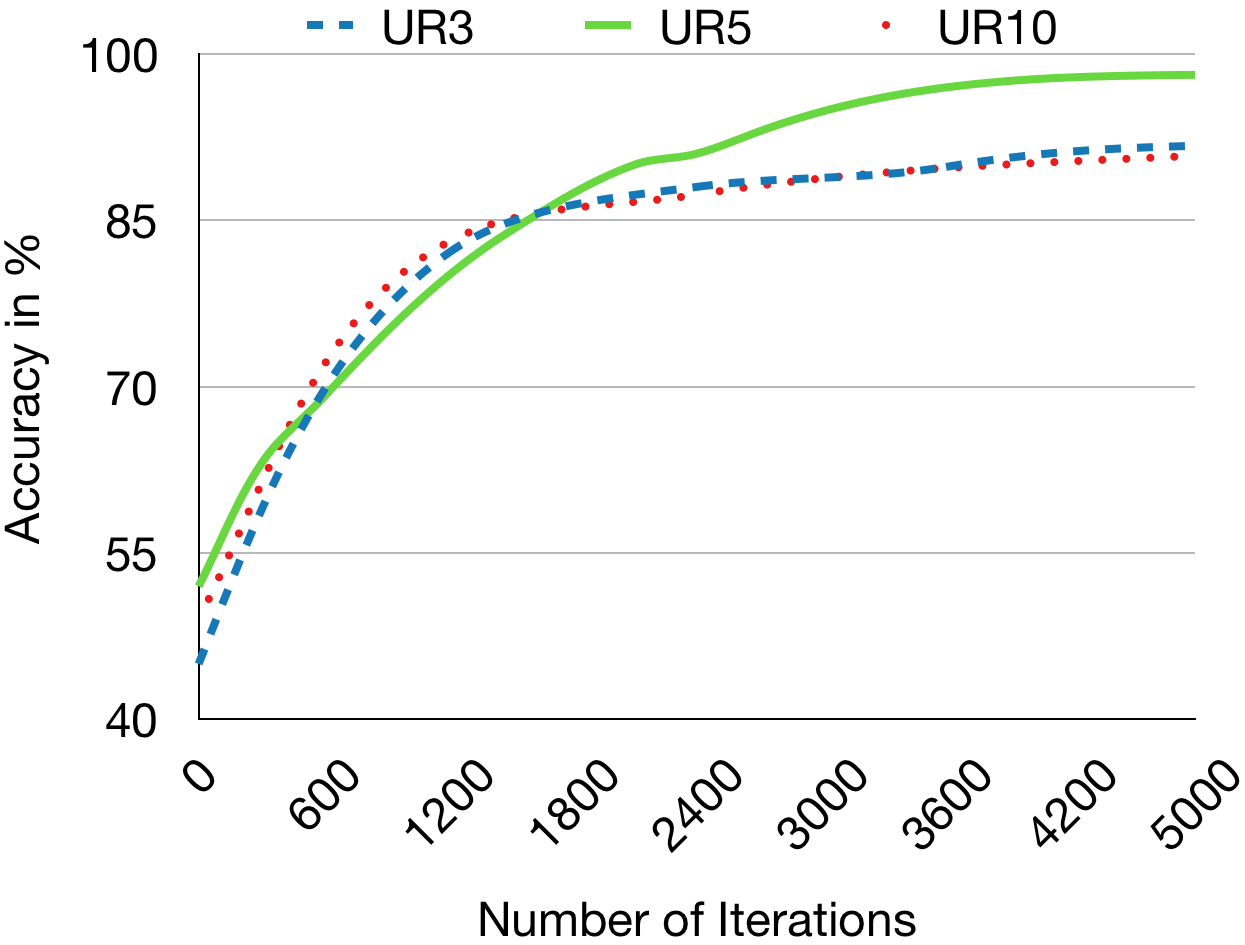}
        \caption{Results: evaluation of the CNN trained for the robot body classification, mask accuracy over a number of iterations for all three robots using validation sets. It can be seen that UR5 outperformed UR3 and UR10.}
        \label{fig:results_mask_accuracy} 
    \end{subfigure}
    \hfill
    \begin{subfigure}[t]{0.325\textwidth}
        \centering
        \includegraphics[width=\linewidth]{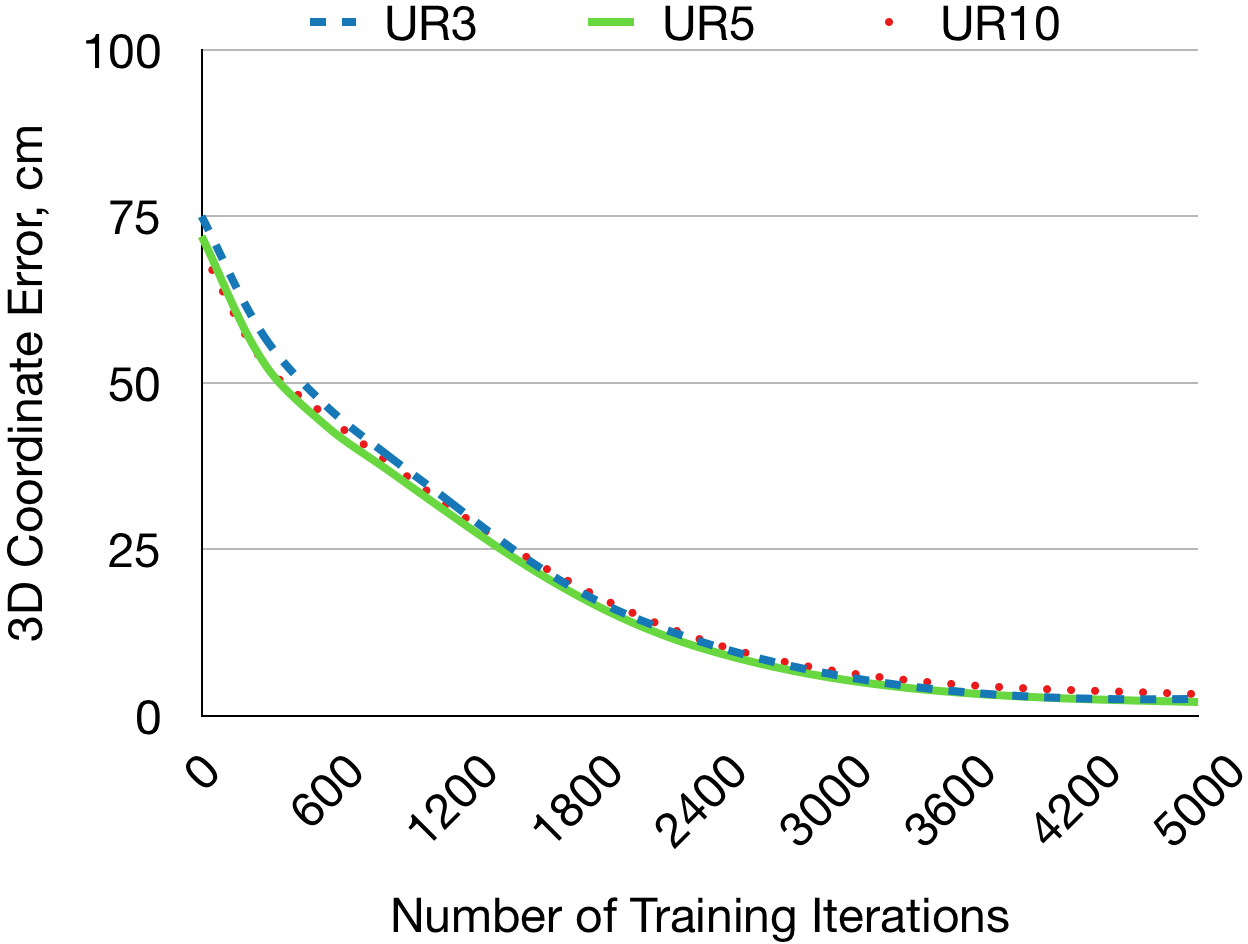}
        \caption{Results: evaluation of the CNN trained for 3D coordinate estimation of the robot joint positions using input based on ground truth mask data.}
        \label{fig:results_coords_accuracy}
    \end{subfigure}
    \begin{subfigure}[t]{0.325\textwidth}
        \centering
        \includegraphics[width=\linewidth]{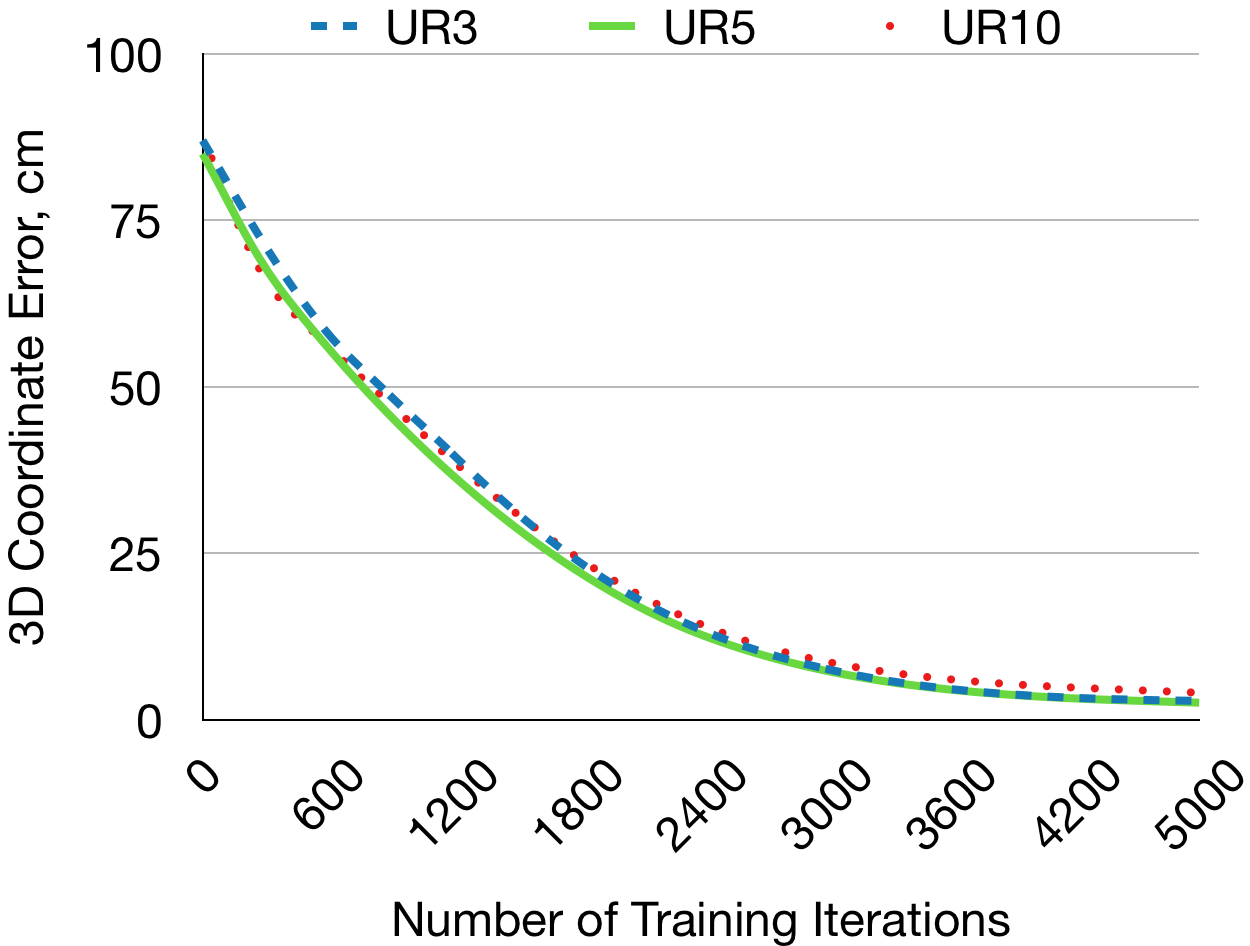}
        \caption{Results: evaluation of the CNN trained for 3D coordinate estimation of the robot joint positions using the full system.}
        \label{fig:results_coords_full_accuracy}
    \end{subfigure}
    
    \caption{Evaluation of our method based on accuracy over a number of training iterations.}
\end{figure*}

\section{CNN TRAINING}

There are two main possibilities on how to train the cascaded CNN. The first option is to train the whole network end-to-end and observe the middle layer of the mask. However, this might not result in the output that is expected and is unlikely to reach the desired mask accuracy. In this work, we train each of the CNNs separately optimising for the best result at each stage. This approach provides the flexibility of using just a part of the system, for example, if only a mask for the robot body is needed. When running the full cascade, the output of the first CNN is used to mask a color input image and use it as an input for the second CNN. The training has been done on each type of the robot separately, however, by observing intermediate-level feature maps, we have noticed very similar features for all of the robot models.

The training of the classification CNN took slightly more than 2 days on a regular nVidia GeForce 1080 GTX graphics card for all the datasets. The data was selected in a random order to avoid any biases and split in mini-batches of $128$ images each for input to avoid overloading memory of the GPU. The learning rate was gradually decreasing, starting at $0.001$ and reducing to $0.000001$ throughout the learning process. It took $5000$ epochs for the network to converge. The input size was half of the original image size: $256\times212$ pixels. The pixel intensity values were converted to float and normalised to lay between 0 and 1. Additionally, pixel values of the ground truth image are clipped to avoid division by zero in cases when the estimated mask fits the ground truth perfectly.

Training of the coordinate estimation CNN was significantly faster, taking under $7$ hours, also converging after $5000$ epochs. The learning rate was adjusted during the training, starting at $0.03$ down to $0.0001$, and momentum was increased over epochs from $0.9$ to $0.999$.

\begin{figure}[ht]
    \includegraphics[width=0.48\textwidth]{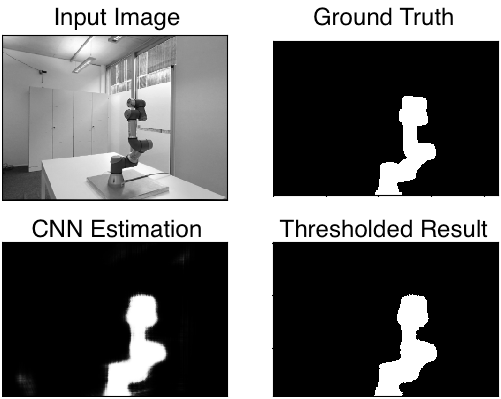}
    \caption{An example result of the UR3 robot body mask classification including input, ground truth, raw and thresholded CNN output images. It can be seen that the mask fit corresponds well with the ground truth. The only drawback is that the fit is not as sharp as the ground truth image. However, no unwanted artefacts or false positives are present.}
    \label{fig:classification_result}
\end{figure}


\section{RESULTS}
\label{sec:results}


For evaluation, we use test sets and analyze outputs of the trained systems against the ground truth data and calculate the accuracy of the system. For the robot body classification, the accuracy is defined by comparing how many pixels in the CNN output mask image match the ground truth mask. For the robot joint coordinates estimation, the error is defined by the Euclidean distance between the estimated coordinates and ground truth in all three dimensions, averaged over all joints of the robot.

First, the results are presented for each of the CNNs separately and then of the whole system altogether. Results are analysed separately for the three types of robots.

\subsection{Evaluation of the Robot Classification Task}

First, we present the results of the robot classification task for each type of the robot. Accuracy is defined by the number of correctly classified pixels in the mask image. Classification of UR5 reached the accuracy of $98,1\%$ and outperformed UR3 and UR10 with $93,1\%$ and $92,8\%$ respectively. The accuracy results over the training iterations can be seen in Figure~\ref{fig:results_mask_accuracy}. An example mask estimation is shown in Figure~\ref{fig:classification_result}.

\begin{figure*}[ht]
\vspace{0.2cm}
    \centering
    \includegraphics[width=0.99\linewidth]{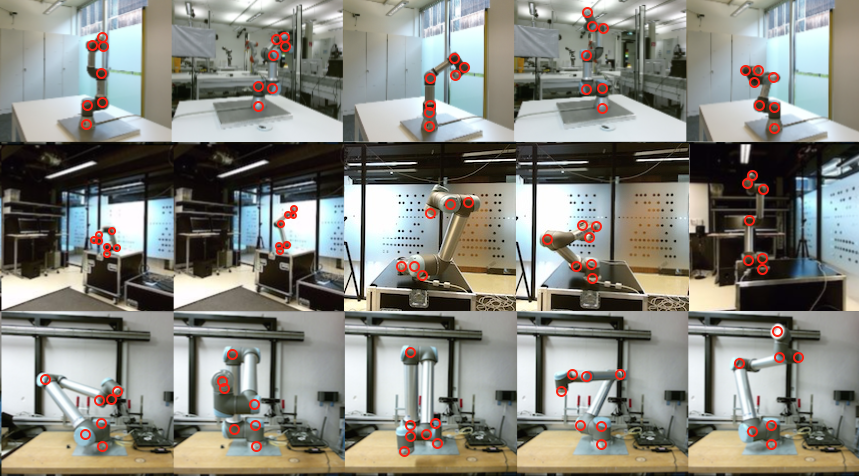}
    \caption{Estimated robot joint position coordinates marked on the images taken from the dataset. Due to difficulty in visualising 3D coordinates on printed figures, the estimated joint coordinates were mapped back into 2D images. Each row represents UR3, UR5 and UR10 robots respectively.}
    \label{fig:marked_coordinates}
\end{figure*}

\subsection{Evaluation of the Robot Joint Coordinate Estimation}
\label{ssec:eval_jointCoords}

In order to analyse the coordinate estimation, first, we use the overlay images based ground truth mask data for the input. As expected, CNN trained on UR5 data provided the most accurate estimation with the average position error of $2,0 cm$, while UR3 had the error of $2,5 cm$ and UR10 - $3,2 cm$. The coordinate estimation results over the training iterations can be seen in Figure~\ref{fig:results_coords_accuracy}.

\subsection{Evaluation of the Full System}

For the full system evaluation, the process is the combination of the previous two methods joined together: the resulting output image of the robot mask classification is used to overlay the color image and passed as an input for the robot joint coordinate estimation. It is imperfect compared to the ground truth data, so worse results were expected compared to the previous test. For the full system, the final coordinate estimation error increased to $2,4 cm$ for UR5, $2,6 cm$ for UR3 and $3,9 cm$ for UR10. Results can be directly compared with the previous Section~\ref{ssec:eval_jointCoords}.

The final results are summarised in Table~\ref{table:results_summary} and the estimated coordinates by the full system marked over the dataset images can be seen in Figure~\ref{fig:marked_coordinates}. Because it is difficult to show 3D estimations on 2D figures, the visualisation of estimation is done by mapping the estimated 3D coordinates back onto input images.


\begin{table}[h]
\caption{Experiment Results Summary}
\label{table:results_summary}
\centering
\begin{tabular}{ |p{3.7cm}||p{1cm}|p{1cm}|p{1cm}|}
 \hline
  & UR3 & UR5 & UR10 \\
 \hline
 Mask Accuracy, \% & $93,1\%$ & $98,1\%$ & $92,8\%$ \\
 Coordinates Error (separate) & $2,5 cm$ & $2,02 cm$ & $3,21 cm$ \\
 Coordinates Error (full system) & $2,57 cm$ & $2,42 cm$ & $3,89 cm$ \\
 \hline
\end{tabular}
\end{table}

\section{CONCLUSIONS AND FUTURE WORK}
\label{sec:conclusion}

In this work, we have addressed robots for collaboration and human-robot interaction tasks. We have found an alternative solution for Hand-Eye calibration and added the flexibility of placing the camera at arbitrary position observing the robot workspace, while still being able to identify the robot in the image and estimate its position.

Our system uses a cascaded convolutional neural network to achieve the goal. For training and testing purposes, we have collected a number of datasets using a line of robots produced by Universal Robots: UR3, UR5 and UR10. This allowed us to precisely train the CNN and achieve the accuracy in robot body classification of up to $98\%$ on the test set and 3D joint coordinate estimation with an error of less than $3 cm$. Furthermore, we have shown that the accuracy directly correlates with the training duration and a number of collected samples. This result is still not accurate enough for applications like visual servoing, but it can be good enough for some collaboration tasks as well as safety alerts in cases where a person does not have to work in a very close proximity to the robot.

Some example applications would be a small body-mounted camera for doctors working in robotised operating rooms or visitors on the factory floor. It would also be useful in robot-robot interaction cases, when a mobile robot is operating in the same areas as the robot arms, either in order to avoid each other, or support the operations by bringing objects, which would be handled by robot manipulators. The system would be trained to identify all the robots existing in the specific environment, and the person would be warned by a visual or audible alert in cases where he gets within the reachable distance of the robot. Furthermore, if the robot gets too close to the person, an emergency stop could be initiated.

For the human-robot collaboration tasks, hand tracking of a person can be achieved using devices like Leap Motion or skeleton tracking to get an estimate of the relative hand positions to the robot. This makes it possible to achieve the tasks like tool handover between the person and the robot, completing joint tasks or even hand-gesture control, while avoiding any unwanted physical contact between the two.

Further work includes expanding our method to new types of robots by using transfer learning from pre-trained CNN. This could allow achieving good accuracy with a limited number of training samples. Furthermore, we will add state-of-the-art skeleton tracking and human motion prediction to perform collaborative human-robot tasks and evaluate the performance compared to the cases of having fixed camera-robot setups.

\section*{ACKNOWLEDGMENT}
This work is partially supported by The Research Council of Norway as a part of the Engineering Predictability with Embodied Cognition (EPEC) project, under grant agreement 240862, and by the Austrian Ministry for Transport, Innovation and Technology (BMVIT) within the project framework CollRob (Collaborative Robotics).

\addtolength{\textheight}{-12cm}   






\bibliographystyle{IEEEtran}
\bibliography{IEEEexample}

\begin{thebibliography}{10}
\providecommand{\url}[1]{#1}
\csname url@rmstyle\endcsname
\providecommand{\newblock}{\relax}
\providecommand{\bibinfo}[2]{#2}
\providecommand\BIBentrySTDinterwordspacing{\spaceskip=0pt\relax}
\providecommand\BIBentryALTinterwordstretchfactor{4}
\providecommand\BIBentryALTinterwordspacing{\spaceskip=\fontdimen2\font plus
\BIBentryALTinterwordstretchfactor\fontdimen3\font minus
  \fontdimen4\font\relax}
\providecommand\BIBforeignlanguage[2]{{%
\expandafter\ifx\csname l@#1\endcsname\relax
\typeout{** WARNING: IEEEtran.bst: No hyphenation pattern has been}%
\typeout{** loaded for the language `#1'. Using the pattern for}%
\typeout{** the default language instead.}%
\else
\language=\csname l@#1\endcsname
\fi
#2}}

\bibitem{roach1987coordinating}
J.~Roach and M.~Boaz, ``Coordinating the motions of robot arms in a common
  workspace,'' \emph{IEEE Journal on Robotics and Automation}, vol.~3, no.~5,
  pp. 437--444, 1987.

\bibitem{leitner2012transferring}
J.~Leitner, S.~Harding, M.~Frank, A.~F{\"o}rster, and J.~Schmidhuber,
  ``Transferring spatial perception between robots operating in a shared
  workspace,'' in \emph{2012 IEEE/RSJ International Conference on Intelligent
  Robots and Systems}.\hskip 1em plus 0.5em minus 0.4em\relax IEEE, 2012, pp.
  1507--1512.

\bibitem{lee2015cyber}
J.~Lee, B.~Bagheri, and H.-A. Kao, ``A cyber-physical systems architecture for
  industry 4.0-based manufacturing systems,'' \emph{Manufacturing Letters},
  vol.~3, pp. 18--23, 2015.

\bibitem{fitzgerald2013developing}
C.~Fitzgerald, ``{Developing Baxter},'' in \emph{Technologies for Practical
  Robot Applications (TePRA), 2013 IEEE International Conference on}.\hskip 1em
  plus 0.5em minus 0.4em\relax IEEE, 2013, pp. 1--6.

\bibitem{miseikis2016automatic}
J.~Miseikis, K.~Glette, O.~J. Elle, and J.~Torresen, ``Automatic calibration of
  a robot manipulator and multi 3d camera system,'' in \emph{System Integration
  (SII), 2016 IEEE/SICE International Symposium on}.\hskip 1em plus 0.5em minus
  0.4em\relax IEEE, 2016, pp. 735--741.

\bibitem{flandin2000eye}
G.~Flandin, F.~Chaumette, and E.~Marchand, ``Eye-in-hand/eye-to-hand
  cooperation for visual servoing,'' in \emph{Robotics and Automation, 2000.
  Proceedings. ICRA'00. IEEE International Conference on}, vol.~3.\hskip 1em
  plus 0.5em minus 0.4em\relax IEEE, 2000, pp. 2741--2746.

\bibitem{schneegans2007using}
S.~Schneegans, P.~Vorst, and A.~Zell, ``{Using RFID Snapshots for Mobile Robot
  Self-Localization.}'' in \emph{EMCR}, 2007.

\bibitem{gutmann1996amos}
J.-S. Gutmann and C.~Schlegel, ``Amos: Comparison of scan matching approaches
  for self-localization in indoor environments,'' in \emph{Advanced Mobile
  Robot, 1996., Proceedings of the First Euromicro Workshop on}.\hskip 1em plus
  0.5em minus 0.4em\relax IEEE, 1996, pp. 61--67.

\bibitem{stasse2008real}
O.~Stasse, A.~Escande, N.~Mansard, S.~Miossec, P.~Evrard, and A.~Kheddar,
  ``Real-time (self)-collision avoidance task on a hrp-2 humanoid robot,'' in
  \emph{Robotics and Automation, 2008. ICRA 2008. IEEE International Conference
  on}.\hskip 1em plus 0.5em minus 0.4em\relax IEEE, 2008, pp. 3200--3205.

\bibitem{de2007skeleton}
A.~De~Santis, A.~Albu-Schaffer, C.~Ott, B.~Siciliano, and G.~Hirzinger, ``The
  skeleton algorithm for self-collision avoidance of a humanoid manipulator,''
  in \emph{Advanced intelligent mechatronics, 2007 IEEE/ASME international
  conference on}.\hskip 1em plus 0.5em minus 0.4em\relax IEEE, 2007, pp. 1--6.

\bibitem{wilson1996relative}
W.~J. Wilson, C.~W. Hulls, and G.~S. Bell, ``Relative end-effector control
  using cartesian position based visual servoing,'' \emph{IEEE Transactions on
  Robotics and Automation}, vol.~12, no.~5, pp. 684--696, 1996.

\bibitem{ruf1997visual}
A.~Ruf, M.~Tonko, R.~Horaud, and H.-H. Nagel, ``Visual tracking of an
  end-effector by adaptive kinematic prediction,'' in \emph{Intelligent Robots
  and Systems, 1997. IROS'97., Proceedings of the 1997 IEEE/RSJ International
  Conference on}, vol.~2.\hskip 1em plus 0.5em minus 0.4em\relax IEEE, 1997,
  pp. 893--899.

\bibitem{daachi2006neural}
B.~Daachi and A.~Benallegue, ``A neural network adaptive controller for
  end-effector tracking of redundant robot manipulators,'' \emph{Journal of
  Intelligent \& Robotic Systems}, vol.~46, no.~3, pp. 245--262, 2006.

\bibitem{siradjuddin2014image}
I.~Siradjuddin, L.~Behera, T.~M. McGinnity, and S.~Coleman, ``{Image-based
  visual servoing of a 7-DOF robot manipulator using an adaptive distributed
  fuzzy PD controller},'' \emph{IEEE/ASME Transactions on Mechatronics},
  vol.~19, no.~2, pp. 512--523, 2014.

\bibitem{soltanpour2013particle}
M.~R. Soltanpour and M.~H. Khooban, ``A particle swarm optimization approach
  for fuzzy sliding mode control for tracking the robot manipulator,''
  \emph{Nonlinear Dynamics}, vol.~74, no. 1-2, pp. 467--478, 2013.

\bibitem{michalos2015design}
G.~Michalos, S.~Makris, P.~Tsarouchi, T.~Guasch, D.~Kontovrakis, and
  G.~Chryssolouris, ``Design considerations for safe human-robot collaborative
  workplaces,'' \emph{Procedia CIrP}, vol.~37, pp. 248--253, 2015.

\bibitem{sheridan2016human}
T.~B. Sheridan, ``Human--robot interaction: status and challenges,''
  \emph{Human factors}, vol.~58, no.~4, pp. 525--532, 2016.

\bibitem{brijacak2017}
I.~Brijacak, S.~Yahyanejad, B.~Reiterer, and M.~Hofbaur, ``Toward safe
  perception in human- robot interaction,'' in \emph{OAGM and ARW Joint
  Workshop, Vienna}.\hskip 1em plus 0.5em minus 0.4em\relax IEEE, May 2017, pp.
  80--85.

\bibitem{yamada1997human}
Y.~Yamada, Y.~Hirasawa, S.~Huang, Y.~Umetani, and K.~Suita, ``Human-robot
  contact in the safeguarding space,'' \emph{IEEE/ASME transactions on
  mechatronics}, vol.~2, no.~4, pp. 230--236, 1997.

\bibitem{mainprice2013human}
J.~Mainprice and D.~Berenson, ``Human-robot collaborative manipulation planning
  using early prediction of human motion,'' in \emph{Intelligent Robots and
  Systems (IROS), 2013 IEEE/RSJ International Conference on}.\hskip 1em plus
  0.5em minus 0.4em\relax IEEE, 2013, pp. 299--306.

\bibitem{mivseikis2016multi}
J.~Mi{\v{s}}eikis, K.~Glette, O.~J. Elle, and J.~Torresen, ``{Multi 3D camera
  mapping for predictive and reflexive robot manipulator trajectory
  estimation},'' in \emph{Computational Intelligence (SSCI), 2016 IEEE
  Symposium Series on}.\hskip 1em plus 0.5em minus 0.4em\relax IEEE, 2016, pp.
  1--8.

\bibitem{simard2003best}
P.~Y. Simard, D.~Steinkraus, J.~C. Platt, \emph{et~al.}, ``Best practices for
  convolutional neural networks applied to visual document analysis.'' in
  \emph{ICDAR}, vol.~3, 2003, pp. 958--962.

\bibitem{Fankhauser2015KinectV2ForMobileRobotNavigation}
P.~Fankhauser, M.~Bloesch, D.~Rodriguez, , R.~Kaestner, M.~Hutter, and
  R.~Siegwart, ``Kinect v2 for mobile robot navigation: Evaluation and
  modeling,'' in \emph{IEEE International Conference on Advanced Robotics
  (ICAR) (submitted)}, 2015.

\bibitem{quigley2009ros}
M.~Quigley, K.~Conley, B.~Gerkey, J.~Faust, T.~Foote, J.~Leibs, R.~Wheeler, and
  A.~Y. Ng, ``{ROS: an open-source Robot Operating System},'' in \emph{ICRA
  workshop on open source software}, vol.~3, no. 3.2, 2009, p.~5.

\bibitem{heikkila2000flexible}
T.~Heikkil{\"a}, M.~Sallinen, T.~Matsushita, and F.~Tomita, ``Flexible hand-eye
  calibration for multi-camera systems,'' in \emph{Intelligent Robots and
  Systems, 2000.(IROS 2000). Proceedings. 2000 IEEE/RSJ International
  Conference on}, vol.~3.\hskip 1em plus 0.5em minus 0.4em\relax IEEE, 2000,
  pp. 2292--2297.

\bibitem{sucan2013moveit}
I.~A. Sucan and S.~Chitta, ``{MoveIt!}'' \emph{Online Available:
  http://moveit.ros.org}, 2013.

\bibitem{meeussen2012urdf}
W.~Meeussen, J.~Hsu, and R.~Diankov, ``{URDF-Unified Robot Description
  Format},'' 2012.

\end{thebibliography}

\end{document}